\documentclass[a4paper]{article}
\usepackage{hyperref}
\usepackage{multirow}
\usepackage{tipa}
\usepackage[T1]{fontenc}
\usepackage[utf8]{inputenc}
\usepackage[svgnames]{xcolor}
\usepackage{xcolor,soul}
\sethlcolor{lightgray}

\usepackage{INTERSPEECH2022}

\title{ASR2K: Speech Recognition for Around 2000 Languages without Audio}

\name{Xinjian Li, Florian Metze, David R Mortensen, Alan W Black, Shinji Watanabe}

\address{
  Carnegie Mellon University}
\email{xinjianl@cs.cmu.edu}

\begin{document}

\maketitle
\begin{abstract}
Most recent speech recognition models rely on large supervised datasets, which are unavailable for many low-resource languages. In this work, we present a speech recognition pipeline that does not require any audio for the target language. The only assumption is that we have access to raw text datasets or a set of n-gram statistics. Our speech pipeline consists of three components: acoustic, pronunciation, and language models. Unlike the standard pipeline, our acoustic and pronunciation models use multilingual models without any supervision. The language model is built using n-gram statistics or the raw text dataset. We build speech recognition for 1909 languages by combining it with Cr\'ubad\'an: a large endangered languages n-gram database. Furthermore, we test our approach on 129 languages across two datasets: Common Voice and CMU Wilderness dataset. We achieve 50\% CER and 74\% WER on the Wilderness dataset with Cr\'ubad\'an statistics only and improve them to 45\% CER and 69\% WER when using 10000 raw text utterances. 



\end{abstract}
\noindent\textbf{Index Terms}: low-resource speech recognition, multilingual speech recognition, endangered languages

\section{Introduction}

Recently, the performance of speech recognition has witnessed rapid improvement due to modern architectures~\cite{gulati2020conformer,karita2019comparative,watanabe2018espnet}. Those models typically require thousands of hours of training data for the target language. However, there are around 8000 languages in the world~\cite{ethnologue}, the majority of which do not have any audio or text datasets. There have been some attempts to reduce the size of the training set by using pretrained features from self-supervised learning models~\cite{baevski2020wav2vec,hsu2021hubert}. However, such models still rely on a small amount of paired supervised data for word recognition. More recently, inspired by the recent success of unsupervised machine translation~\cite{conneau2017word,artetxe2018robust}, there is some work applying the unsupervised approach to speech recognition as well~\cite{baevski2021unsupervised}. Those models apply adversarial learning to automatically learn a mapping between audio representations and phoneme units. They can learn a phoneme recognition model using an unlabeled audio dataset and a text dataset. 

Despite the success of those recent approaches, all of these models rely on some audio datasets of the target language (labeled or unlabeled), which significantly restricts the scope of target languages. In this work, we investigate whether we can develop speech recognition systems without requiring any audio dataset or pronunciation lexicon for the target language. The only assumption is the existence of some monolingual text or a set of n-gram statistics for the target language. Our proposed method consists of three components: acoustic, pronunciation, and language models. Both acoustic and pronunciation models can be trained using supervised datasets from high-resource languages, and then applied to the target language by taking advantage of some linguistic knowledge. Both models can be applied in a zero-shot learning fashion without any supervision. Finally, we use the raw texts or n-gram statistics to create a language model, which is then combined with the pronunciation model to create a WFST decoder. To analyze our pipeline more efficiently with small test sets, we also propose an approach to decompose the observed errors into acoustic/pronunciation model errors and language model errors.


We apply our approach to 1909 languages using Cr\'ubad\'an: a large endangered languages n-gram database and then test our approach on 129 languages: 34 languages from the Common Voice and 95 languages from CMU Wilderness dataset~\cite{ardila2019common,black2019cmu}. On the Wilderness dataset, we achieve 50\% CER (character error rate) and 74\% WER  (word error rate) respectively when using Cr\'ubad\'an's statistics only, and improve them to 45\% CER and 69\% WER by using 10000 raw text utterances. As far as we know, this is the first attempt to build speech recognition for thousands of languages without audio.





\section{Related Work}

Most speech recognition approaches can be classified into one of several groups depending on their data requirements. The most common group has access to the paired supervised dataset $\mathcal{D} = \{ (X_i, Y_i) \}^N_{i=1}$ where $(X, Y)$ is a paired audio and text of an utterance. If the size $N$ of the dataset is large enough, various end-to-end models can be trained using CTC, ASG, seq2seq, RNN Transducer, and other objectives~\cite{graves2006connectionist, collobert2016wav2letter, graves2013speech, sutskever2014sequence}. If the size is small, then it would be a low-resource speech recognition in which some acoustic knowledge should be transferred from high-resource languages~\cite{li2019multilingual,xu2020lrspeech}. Self-supervised training takes advantage of another large raw speech dataset $\{ X_j \}$ to learn hidden representations of speech signals, those representations are useful to the supervised tasks and can reduce the amount of the paired dataset~\cite{baevski2020wav2vec,hsu2021hubert}. The semi-supervised learning approach also leverages unlabeled speech datasets or text datasets to augment the supervision set~\cite{vesely17_interspeech,synnaeve2019end,rosenberg2019speech}. 


Recently, unsupervised speech recognition attempts to target the dataset $\mathcal{D} = (\{X_i \}_{i=1}^I, \{Y_j \}_{j=1}^J)$ where we have access to an unlabeled raw audio set $\{X_i \}_{i=1}^I$ and a raw text dataset $\{Y_j \}_{j=1}^J$~\cite{baevski2021unsupervised}. The audio and text do not need to be aligned with each other. A generator model is jointly trained with a discriminator model. The generator model attempts to translate audio into phonemes, while the discriminator model attempts to distinguish between phonemes transliterated from text and phonemes recognized from the generator. The disadvantage of this direction is that the model could only recognize phonemes instead of words and it requires a phonemizer (pronunciation model) for the target language, which would not be available for most languages. Another related direction is unsupervised speech unit discovery~\cite{chorowski2019unsupervised,tjandra2019vqvae}, which is similar to the self-supervised learning approach and attempts to discover phone units from audios $\mathcal{D}= \{ X_i \}_{i=1}^I$. This group of approaches, however, cannot emit explicit phonemes or words as it does not have knowledge of the lexicon and language model for the target language.

In this work, we propose a new paradigm to focus on the text-only dataset $\mathcal{D} =\{Y_j \}_{j=1}^J$. While all the previous groups require some amount of audio datasets $\{ X_i \}$ (paired or unpaired) for the word recognition of the target language, we argue this requirement can be relaxed to some extent.

\section{Model}


Our speech pipeline is divided into the \textit{acoustic model}, \textit{pronunciation model} and \textit{language model}. The joint probability over speech audio $X$ and speech text $Y$ can be factorized as

\begin{equation}
    p_\theta(X,Y) = \sum_P p_{\text{am}}(X | P) p_{\text{pm}}(P | Y) p_{\text{lm}}(Y)
\end{equation}
, where $P$ is the phoneme sequence corresponding to the text $Y$. The pronunciation model $p_{\text{pm}}$ is typically modeled as a deterministic function $\delta_{\text{pm}}$. In our pipeline, only the language model can be estimated from the text, both the acoustic model and pronunciation model are approximated using zero-shot learning or transfer learning from other high resource languages, therefore we denote $\hat{p}_{\text{am}}, \hat{\delta}_{\text{pm}}$ for the approximated acoustic model and pronunciation model. The previous factorization can be approximated by

\begin{equation}
    p_\theta(X,Y) \approx \hat{p}_{\text{am}}(X | \hat{P}) p_{\text{lm}}(Y)
\end{equation}
, where $\hat{P}= \hat{\delta}_{\text{pm}}(Y)$ is the approximated phonemes. 

\subsection{Acoustic Model}

The acoustic model should be able to recognize phonemes of the target languages even when the languages are unseen in the training set. We follow a direction of recently proposed allophone-based multilingual architectures~\cite{li2020universal,li2021hierarchical}. This direction attempts to recognize phonemes of an unseen language using language-independent phone representations and their mappings to the language-dependent phonemes. Essentially, those architectures attempt to represent the acoustic model as follows:


\begin{equation}
    \hat{p}_{\text{am}}(P|X) = \sum_Q p_{\text{lang}}(P|Q)p_{\text{uni}}(Q|X)
\end{equation}
, where $p_{\text{uni}}(Q|X)$ is a language-independent universal phone recognition model, recognizing physical-level phone units $Q$ from the speech audio $X$. The allophone architecture $p_{\text{lang}}(P|Q)$ is to encode how each physical phone should be mapped to a language-dependent phoneme. The relation between phones and phonemes is called an \textit{allophone}, which is usually encoded as a $1$-$n$ deterministic function annotated by phonologists for each language. The mapping is easier to obtain than the supervised dataset for low resource languages. We rely on Allovera and PHOIBLE datasets for allophone mapping of more than 2000 languages~\cite{mortensen2020allovera,phoible}.  The other model $p_{\text{uni}}(Q|X)$ does not have any dependency on the target language, therefore it can be trained using high-resource languages such as English and Mandarin. The CTC objective is used to train this acoustic model~\cite{graves2006connectionist}. The conditional independence assumption in CTC prevents the model from biasing too much towards one specific language model (e.g: English), therefore it can be easier to apply to other low-resource languages. In our experiment, we observe the originally proposed model~\cite{li2021hierarchical}, is not very robust when recognizing audios from different domains. To further improve the model, instead of using the standard filterbank features, we use self-supervised learning (SSL) features as our frontend feature extraction~\cite{baevski2020wav2vec,hsu2021hubert,conneau2020unsupervised}.




\subsection{Pronunciation Model}

The pronunciation model is essentially a G2P (grapheme-to-phoneme) model that can predict the phoneme pronunciation given a grapheme sequence: $P = \delta_{\text{pm}}(Y)$. For high-resource languages, the G2P model can be either trained using a dictionary or be developed using rule-based systems~\cite{cmudict,mortensen2018epitran}. However, the majority of the languages do not have any accessible dictionaries or rules, therefore we consider an approximated pronunciation model $\hat{\delta}_{\text{pm}}$ instead. In particular, we apply a recently proposed multilingual G2P model as our pronunciation model~\cite{li2022g2p}. For any target language $l_{\text{target}}$, this G2P model selects top-$k$ nearest languages: $l_{\text{topk}} \in \text{KNN}(l_{\text{target}})$ whose training set is available, then during the inference, it first propose $k$ hypothesis using each nearest language model $\delta_{l_{\text{topk}}}$, the models are ensembled by combining hypothesis into a lattice to emit the most-likely approximated sequence:

\begin{equation}
    \hat{\delta}_{l_{\text{target}}} = \text{Ensemble}(\{ \delta_{l_{\text{topk}}} | l_{\text{topk}} \in \text{KNN}(l_{\text{target}}) \})
\end{equation}
The similarity metric between languages is defined to be the shortest path of two languages on the phylogenetic tree (i.e: language family tree). This approach enables us to approximate the pronunciation model for every language in Glottolog database~\cite{nordhoff2011glottolog}, which contains phylogenetic information about 7915 languages.

\subsection{Language Model}

For the language model, we first estimate the vocabulary $V = \{ w_1, w_2, ..., w_{|V|} \}$ from the raw text dataset $\{ Y_i \}$. For each word $w_i \in V$, its pronunciation can be approximated using the pronunciation model and then this lexicon information can be encoded into a lexicon graph $L$. The text dataset also enables us to estimate the classical n-gram language model by counting n-grams statistics $C(w_1, ..., w_n)$. This n-gram language model can be then encoded into a grammar graph $G$. Composing the lexicon graph $L$ and the grammar graph $G$ as well as the CTC topology graph $H$ would generate a WFST-based language decoder $\text{HLG}$~\cite{miao2015eesen}. 

We realize that the text dataset requirement $\{ Y_i \}$ can be further relaxed as the building blocks of the $\text{HLG}$ graph only consist of the statistics $\{ V, C \}$ estimated from the text dataset. For languages whose text dataset $\{ Y_i \}$ is absent but $\{ V,C \}$ is available, we can still proceed to build the decoder $\text{HLG}$. This is common for many languages in the internet: while only a few hundred languages are recognized as being in use for web texts on the World Wide Web~\cite{ethnologue}, there exists several large databases collecting lexicon-related statistics for thousands of languages. For example, Cr\'ubad\'an is a database consisting of vocabulary, bigrams, and character-trigrams statistics for around 2000 languages~\cite{scannell2007}. Employing statistics from it, we build speech recognition systems for around 2000 languages.

\subsection{Error Decomposition}

Since the acoustic, pronunciation models are approximated models, it is helpful to understand how the approximation would impact our results. As the final observed errors also contain the language model errors, we propose a framework to decompose the observed errors $\epsilon_{\text{observed}}$ into language model errors and other errors. To achieve this, in addition to the experiment using the approximated models, we conduct a new set of experiments using the \textit{oracle} acoustic and pronunciation models (i.e. the acoustic and pronunciation model that achieves perfect performance), such that any recognition errors in this new experiment should be attributed to the language model $\epsilon_{\text{lm}}$. The gap between the observed error $\epsilon_{\text{observed}}$ and the oracle error $\epsilon_{\text{lm}}$ should correspond to the errors made by the approximated acoustic and pronunciation model $\epsilon_{\text{am/pm}}$. In other words, the observed errors can be decomposed as follows:


\begin{equation}
    \epsilon_{\text{observed}} = \epsilon_{\text{am/pm}} + \epsilon_{\text{lm}} 
\end{equation}

To estimate the oracle error $\epsilon_{\text{lm}}$, every testing utterance is first converted to the phoneme sequence using our pronunciation model, the phoneme sequence is then augmented with the CTC blank labels by inserting blank labels "$\langle \text{blk} \rangle$" between every pair of phonemes. (e.g: "a b" is converted to "$\langle \text{blk} \rangle$ a $\langle \text{blk} \rangle$ b $\langle \text{blk} \rangle$"). Next, the augmented sequence is converted to CTC logits by giving an extremely high probability to each phoneme (including blank) for every timestep. Finally, the logits is fed into the decoder $\text{HLG}$ to be decoded. We obtain the oracle error $\epsilon_{\text{lm}}$ by comparing it against the expected word sequence. The achieved error rate is the oracle error rate, as we assume the pronunciation model is perfect: pronunciation in logits is perfectly consistent with the pronunciation in the $\text{HLG}$ decoder. The acoustic model is perfect as well: it assigns extremely high probability to the ``correct'' phoneme.

\section{Experiments}


For the acoustic model, we tried 4 different models, one from the previous literature and the newly proposed SSL-based models~\cite{li2021hierarchical}. All the models are trained using cmn, deu, eng, fra, ita, rus, tur, vie languages from the Common Voice dataset~\cite{ardila2019common}. In the SSL-based model, we tried three different self-supervised learning features: HuBERT, wav2vec2, and XLSR~\cite{baevski2020wav2vec,hsu2021hubert,conneau2020unsupervised}. All the features are extracted using s3prl framework~\cite{yang21c_interspeech}. For every SSL model, the features from the last hidden layer were used. Two layers of transformers are appended on top of the pretrained features, which are then connected with the multilingual architecture $p_{\text{lang}}(P|Q)$ as proposed in the original literature~\cite{li2021hierarchical}. The transformer layer has a 768 hidden size and 4 multi-attention heads. Other parameters follow the original literature~\cite{li2021hierarchical}. For the pronunciation model, we use the multilingual model proposed in the previous literature and its implementation~\cite{li2022g2p}~\footnote{\url{https://github.com/xinjli/transphone}}. For the language model,  we first download the complete dataset from Cr\'ubad\'an's website~\cite{scannell2007}, which results in 1909 languages after cleaning. Each language consists of several files: unigrams, bigrams, web urls for the target language, and character trigrams. The most relevant files are unigrams (vocabulary) and bigrams. We provide statistics in Table~\ref{tab:Cr\'ubad\'an}. The same set of information can also be extracted from raw text sets  $\{ Y_i \}$ if we have access to them. For the WFST decoder, we use the k2 library and adapt its icefall recipe\footnote{\url{https://github.com/k2-fsa/icefall}}. We build trigram models from texts and bigram models from Cr\'ubad\'an. During the decoding, we set the search beam size to be 20, output beam size to be 8, min and max active states to be 30 and 10000.


\begin{table}[tbh]
  \centering
    \caption{Descriptive statistics for distinct unigrams and bigram for 1909 languages from Cr\'ubad\'an database. }
  \begin{tabular}{lrrrrr}
    \toprule
     & mean & std & 25\% & median & 75\% \\
    \midrule
    unigram & 10870 & 14012 & 837 & 5149 & 14761 \\ 
    bigram & 29383 & 22087 & 2504 & 42996 & 50000 \\
    \bottomrule
  \end{tabular}
  \label{tab:Cr\'ubad\'an}
\end{table}

To test our approach on unseen languages, we use the 34 languages from Common Voice dataset (denoted by CV) and 95 languages from CMU Wilderness corpus (denoted by WN)~\cite{black2019cmu}. For the Common Voice languages, we select the subset of languages whose dataset is larger than 1000 utterances. Any languages seen in the acoustic model are excluded (i.e: cmn, deu, eng, fra, ita, rus,tur, vie). 95 languages from Wilderness are selected based on the top-100 MCD score, which measures the alignment qualities. 5 languages are excluded due to duplications and preprocess failure (i.e: gag, xsb, nah, may, pxm).



\subsection{Results}

We first evaluate the acoustic model using PER (phoneme error rate). Note that our PER is only an approximation of the actual PER as the expected phoneme sequence relies on the pronunciation model, which is only an approximation. However, it reveals many useful insights into the acoustic models. Table~\ref{tab:am} shows the performance across 4 models. The baseline acoustic model has around 50\% PER and half of the errors are deletion errors (e.g: /a/, /i/ are our most deleted phonemes). We find the main causes of deletion errors are the domain mismatch and language mismatch. To improve the robustness, we employ the SSL-based models, which decreases the error rate by 5\%. Most of the improvement is from the deletion reduction. We find the XLSR model, which is a multilingually pretrained model, performs the best and we use it as the main model in the pipeline.


\begin{table}[tbh]
  \centering
    \caption{Average results (\%) of the acoustic model on all test languages. PER is the  phoneme error rate,  Ins, Del, Sub are Insertion, Deletion and Substitution Error. CV and WN denote Common Voice and Wilderness datasets.}
  \begin{tabular}{lrrrrr}
    \toprule
    Acoustic Model & \multicolumn{2}{c}{PER} & Ins & Del & Sub \\
    & CV & WN \\
    \midrule
    Baseline~\cite{li2021hierarchical} & 51.7 & 49.2 & 1.02 & 30.2 & 19.7 \\
    SSL (HuBERT)~\cite{hsu2021hubert} & 49.7 & 44.3 & 1.15 & 23.8 & 20.8 \\
    SSL (wav2vec2)~\cite{baevski2020wav2vec} & 49.8 & 43.4 & 1.37 & 25.8 & 18.1 \\
    SSL (XLSR)~\cite{conneau2020unsupervised} & \textbf{47.8} & \textbf{42.1} & 1.49 & 24.7 & 19.2\\
    \bottomrule
  \end{tabular}
  \label{tab:am}
\end{table}

\begin{table}[tbh]
  \centering
    \caption{Average Performance (\%) of the language model on all testing languages under different resource conditions. CER, WER denotes character error rate and word error rate.}
  \begin{tabular}{lrrrr}
    \toprule
    Language Model &  \multicolumn{2}{c}{CER} & \multicolumn{2}{c}{WER} \\
     & CV & WN & CV & WN \\
    \midrule
    Cr\'ubad\'an Model & 65.5 & 50.2 & 92.4 & 74.5 \\
    Text Model (1k utterances) & 55.3 & 50.8 & 84.6 & 76.9 \\
    Text Model (5k utterances) & 51.3 & 47.0 & 80.2 & 72.2 \\
    Text Model (10k utterances) & \textbf{50.9} & \textbf{44.9} & \textbf{79.0} & \textbf{69.2} \\
    \bottomrule
  \end{tabular}
  \label{tab:lm}
\end{table}

\begin{table}[tbh]
  \centering
        \caption{A Welsh example from the Common Voice dataset. The top two rows are the hypothesis (HYP) and reference (REF) phonemes, the bottom two rows are the hypothesis and reference words. Deleted phonemes and words are highlighted.}
  \begin{tabular}{ll}
    \toprule
    Model & Sentence \\
    \midrule
    HYP &  k\textopeno p\texttheta\textchi i:\dh\textepsilon rp\textschwa nv\textschwa nk\textschwa sf\textopeno n\textschwa k\textschwa nhar\textschwa\textchi \\
    REF & k\textopeno p\hl{e:a}\texttheta \hl{j\textopeno i:j}\textchi i:\dh\textepsilon rp\textschwa nvi:n\hl{\textepsilon}k\textepsilon sfo:n\textschwa \hl{n}k\textschwa nhara\textchi \\
    \midrule
    HYP & gobeithio ch dderbyn yn gyson cynharach \\
    REF & gobeithio \hl{i} chi dderbyn \hl{fy neges ff\^{o}n} yn gynharach \\
    \bottomrule
  \end{tabular}
  \label{tab:welsh}
\end{table}

Table~\ref{tab:lm} shows the language model performance (using XLSR as the acoustic model). First, we try n-gram statistics from the Cr\'ubad\'an without using any text dataset. It shows that Cr\'ubad\'an captures some character-level information even without any text dataset: it achieves 65\% and 50\% CER on two datasets. The Cr\'ubad\'an WER of the Wilderness languages is also very promising under this condition: 74.5\%. Next, we use 1k, 5k, 10k text utterances from the training set to train the model without Cr\'ubad\'an. As the training text datasets are in the same domain as the test dataset, this improves the performance significantly. With 10k text dataset, we achieve 51\% and 45\% CER respectively. While we omit the result in this table, we also investigate the effect of combining Cr\'ubad\'an and text language models together. However, it does not improve the performance because there is a domain mismatch between two models. The text-only language model shown in Table.\ref{tab:lm} performs the best.

To understand the language model errors, we compute the insertion, deletion, and substitution errors. We find the dominant errors are deletion and substitution. By comparing the most common word errors and phoneme errors, we observe that the phoneme errors have been propagated into the word errors: the previous deletion of phonemes /a/, /i/ caused deletions of the entire words, especially of some short words (e.g: \textit{na}, \textit{ni}). The substitution error also suffers from the missing phonemes issue. For example, the most substituted pair is (\textit{charirca}, \textit{carica}), it is clear that our model failed to recognize several consonants. Table~\ref{tab:welsh} shows a typical example from the pipeline. The acoustic model tends to recognize fewer phones from the audio, those phoneme deletions propagate to the language model and lead to the word deletions.

\subsection{Error Decomposition Analysis}

\begin{figure}[t]
    \centering
    \includegraphics[width=\linewidth]{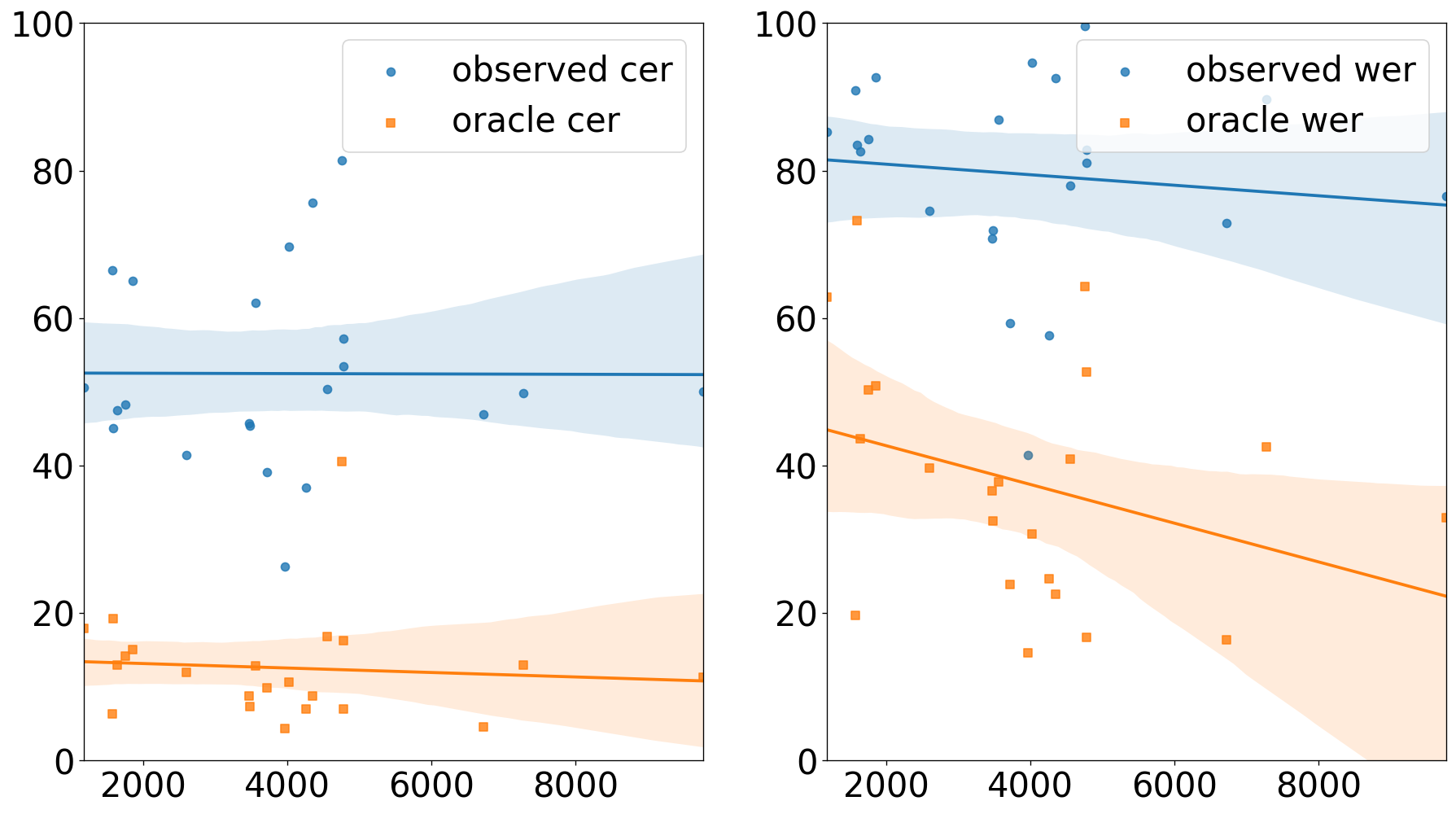}
    \caption{The trend of CER (left) and WER (right) using different sizes of training text. The horizontal axis represents the size of the text dataset. The vertical axis is the error. Each blue circle point denotes an observed error $\epsilon_{\text{observed}}$ from a particular language in the Common Voice corpus and each orange square point shows the oracle error $\epsilon_{\text{lm}}$. An OLS estimator is applied to all sets of points. }
    \label{fig:lowerbound}
\end{figure}

Next, we apply the error decomposition framework to our results in Figure~\ref{fig:lowerbound}. The figure shows the trend of how the CER/WER responds to the size of the training text dataset.  Each blue circle point on the top region represents an observed error $\epsilon_{\text{observed}}$ from the Common Voice corpus and each orange square point on the bottom region is an oracle error $\epsilon_{\text{lm}}$ with our framework. It shows that both errors tend to decrease as the size of the text dataset increases, however, the oracle error has a much sharper decreasing slop than the observed one. As we mentioned in the previous section, the oracle error shows the errors from the language model and the gap between the two errors is the error from the acoustic and pronunciation model. Based on this assumption, the figure indicates that 30 $\sim$ 40\% word errors are from the language model and 40 $\sim$ 50\% word errors are from the acoustic model and pronunciation model; most of the character errors are caused by the acoustic model and pronunciation model.

\subsection{Language Analysis}
    
We can also interpret the results from the linguistic perspective and discuss several limitations of the pipeline. First, we find the phonology of the target language has a crucial impact on the PER performance. Since our acoustic model is trained using high-resource languages (most of them, Indo-European) and then applied to the target language, phonemes that are not common in Indo-European languages should be difficult to recognize. For example, non-pulmonic consonants are common in some languages (e.g: implosive consonants are widespread in Sub-Saharan Africa) but are not typical phonemes in high-resource languages. Another example is the tonal language, we find the Sochiapam Chinantec language displays bad performance: 73\% PER, 75.9\% CER, and 96.5\% WER. This language is a tonal language with 7 different tones. The acoustic model is trained without tonal information and fails to distinguish tonal contrasts (Mandarin Chinese, a tonal language, is included in the acoustic training set, but the tonal information was not used during the training). Orthography depth is another important factor for acoustic performance. The pronunciation model tends to fail more frequently when the language has a deeper orthography (i.e. the rules to map graphemes to phonemes are complicated). For instance, the Swedish language has deep orthography, which makes the PER (67\%) significantly worse than the average PER. Furthermore, if the writing system of the target language is unknown to the pronunciation model, then the model cannot infer its pronunciation. In our dataset, the Maldivian language is written in the Thaana script, which is mostly unknown to the pronunciation model. The error rates are 80\% PER, 81\% CER and 99\% WER. Finally, we observe that some languages have relatively small gaps between CER and WER, and others have larger gaps. For example, the Tai Dam language has an error rate gap of less than 20\%. On the other hand, in the closely related Northern Thai language, we observe a gap of around 40\%. We find the length of a typical word is the main cause: the average token length in Tai Dam is 3.48 characters (e.g., \textit{choi}), but Northern Thai has an average token length of 7.04 (e.g.: \textit{we-machi-warbogwad-e-nandi}). We find there is a strong correlation between the gap and the length of word ($r=0.8015$, in our experiment).

\section{Conclusion}

In this paper, we propose a speech recognition pipeline using raw text or n-gram statistics, and we apply it to around 2000 languages. Our training scripts will be released for more researchers to explore this direction.\footnote{our code will be available at \url{https://github.com/xinjli/asr2k}}

\bibliographystyle{IEEEtran}

\bibliography{mybib}


\end{document}